\documentclass{mva_style}
\usepackage{support-caption}
\usepackage{subcaption}
\usepackage{adjustbox}
\usepackage{booktabs}
\usepackage{multirow}
\usepackage{xurl}

\usepackage{url}

\usepackage[breaklinks=true,bookmarks=true,colorlinks=true]{hyperref}

\finalcopy 

\begin{document}
\title{MVA2023 Small Object Detection Challenge for Spotting Birds:\\ Dataset, Methods, and Results}

\author{
Yuki Kondo$^1$ \and Norimichi Ukita$^2$ \and Takayuki Yamaguchi$^3$ \and
Hao-Yu Hou$^4$ \and Mu-Yi Shen$^4$ \and Chia-Chi Hsu$^4$ \and En-Ming Huang$^4$ \and Yu-Chen Huang$^4$ \and Yu-Cheng Xia$^4$ \and Chien-Yao Wang$^5$ \and  Chun-Yi Lee$^4$ \and
Da Huo$^6$ \and Marc A. Kastner$^7$ \and Tingwei Liu$^6$ \and  Yasutomo Kawanishi$^{8,6}$ \and Takatsugu Hirayama$^{9,6}$ \and Takahiro Komamizu$^6$ \and Ichiro Ide$^6$ \and 
Yosuke Shinya$^{10}$ \and
Xinyao Liu$^{11}$ \and Guang Liang$^{11}$ \and
Syusuke Yasui$^{12}$ \and\\
\normalsize $^1$Toyota Motor Corporation, 
$^2$Toyota Technological Institute,
$^3$Iwate Agricultural Research Center,\\
\normalsize $^4$National Tsing Hua University,
$^5$Institute of Information Science, Academia Sinica,
$^6$Nagoya University,\\
\normalsize $^7$Kyoto University,
$^8$RIKEN,
$^9$University of Human Environments,
$^{10}$Independent Researcher,\\
\normalsize $^{11}$Xi'an Jiaotong University,
$^{12}$Space shift inc. 
}

\maketitle

\section*{\centering Abstract}
\textit{
Small Object Detection (SOD) is an important machine vision topic because (i) a variety of real-world applications require object detection for distant objects and (ii) SOD is a challenging task due to the noisy, blurred, and less-informative image appearances of small objects.
This paper proposes a new SOD dataset consisting of 39,070 images including 137,121 bird instances, which is called the Small Object Detection for Spotting Birds (SOD4SB) dataset.
The detail of the challenge with the SOD4SB dataset~\footnote{Challenge site~\cite{sodbchallenge2023misc}: \url{https://www.mva-org.jp/mva2023/challenge}} is introduced in this paper.
In total, 223 participants joined this challenge.
This paper briefly introduces the award-winning methods.
The dataset~\footnote{Dataset: \url{https://drive.google.com/drive/u/2/folders/1vTHiIelagbzPO795yhOdNUFh9u2XxZP-}}, the baseline code~\footnote{Baseline code~\cite{baselinecode_mva2023_sod_challenge}: \url{https://github.com/IIM-TTIJ/MVA2023SmallObjectDetection4SpottingBirds}}, and the website for evaluation on the public testset~\footnote{Codalab~\cite{codalab_competitions} site: \url{https://codalab.lisn.upsaclay.fr/competitions/9594}\label{footnote:codalab}} are publicly available.
}

{\let\thefootnote\relax\footnotetext{%
\hspace{-5mm} Yuki Kondo, Norimichi Ukita and Takayuki Yamaguchi are the MVA2023 Small Object Detection Challenge for Spotting Birds organizers. The other authors participated in the challenge.\\
Appendix.\ref{sec:apd:team} contains the authors' team names and affiliations.

}
}

\section{Introduction}

\begin{figure}[t]
  \begin{center}
    \includegraphics[width=\linewidth]{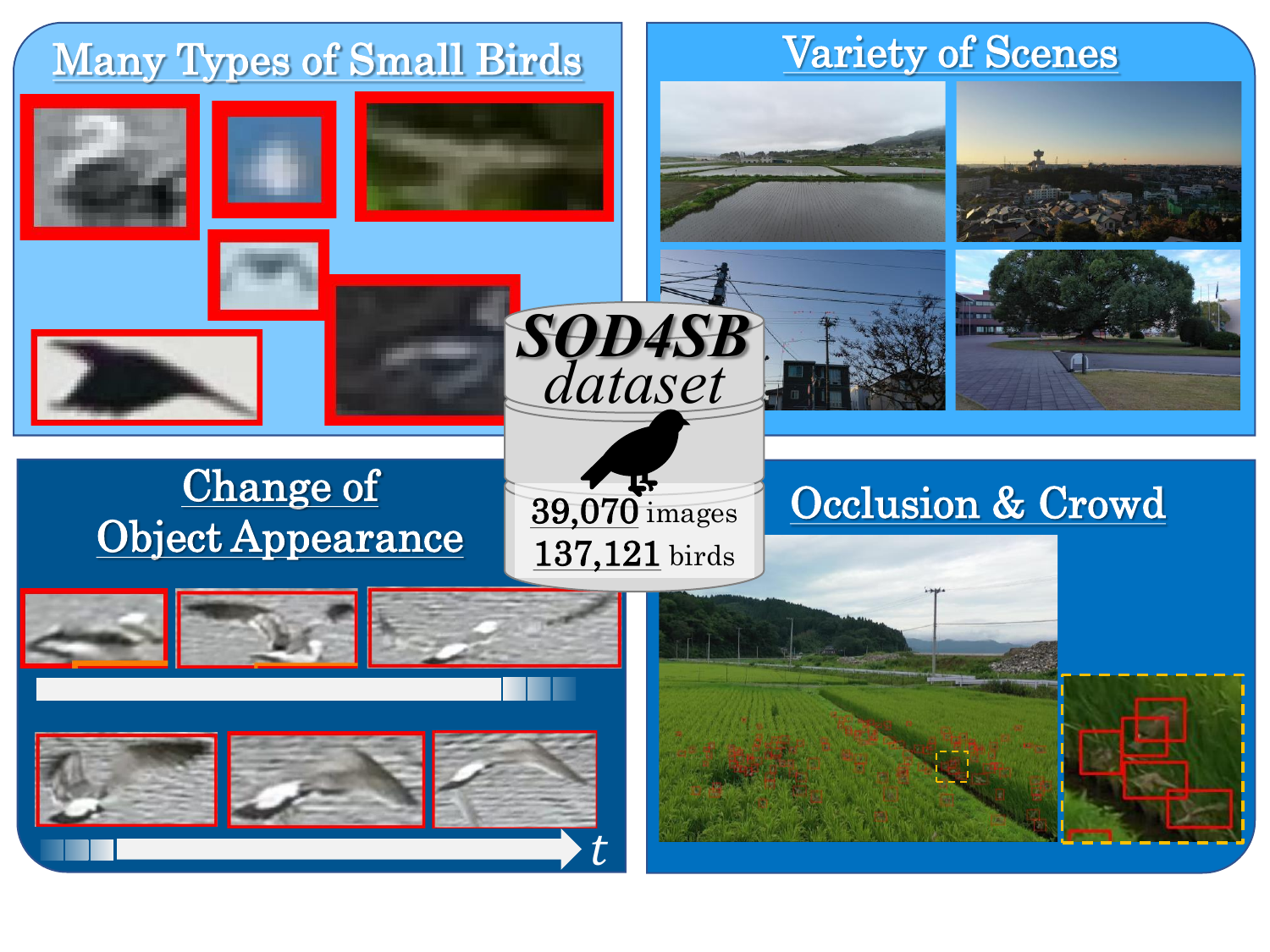}
  \end{center}
  \vspace{-2em}
  \caption{Overview of the Small Object Detection for Spotting Birds (SOD4SB) dataset. The SOD4SD dataset contains a wide variety of small bird types and a variety of scenes. In addition, the flight behavior of the birds and the movements of the UAVs as they film the scene can significantly change bird appearance, and flocking behavior can cause birds to occlude each other, which makes the SOD task even more challenging.}
  \label{fig:overview_sod4sb}
\end{figure}

Object detection is one of the fundamental technologies in the field of machine vision.
Its performance has been improved by Convolutional Neural Networks (CNN)~\cite{Faster_R-CNN_NIPS2015,yolo,duan2019centernet,tan2020efficientdet} and Vision Transformers (ViT)~\cite{detr,wang2023internimage,fang2023eva}.
The performance of the object detection task is evaluated in several huge datasets such as COCO~\cite{COCO_ECCV2014} and PASCAL VOC~\cite{everingham2010pascal}.
Compared with common object detection tasks, SOD~\cite{sod_survey1,sod_survey2,sod_survey3} is still challenging due to the noisy, blurred, and less-informative image appearances of small objects.
One of the reasons of the immatureness of SOD is a limited amount and variety of SOD datasets~\cite{wang2021tiny,yu2020scale,behrendt2017deep,waqas2019isaid,bosquet2018stdnet} and evaluation platforms~\cite{yu20201st,coluccia2021drone}.

Considering the aforementioned issues in SOD, we organized the SOD challenge, {\em Small Object Detection Challenge for Spotting Birds}, with our new dataset of SOD for spotting birds.
Compared with visually-simple objects that are targeted in the previous SOD datasets (e.g., pedestrians~\cite{yu2020scale,zhang2019widerperson} and rigid objects such as vehicles and ships~\cite{waqas2019isaid,behrendt2017deep,wang2021tiny,coluccia2021drone} captured from bird-eye views), wild birds (i) change their moving paces and silhouettes, (ii) freely fly not on the ground plane but three-dimensionally, (iii) are often crowded, and (iv) are observed in front of a variety of background regions (e.g., sky, clouds, trees, mountains, and so on) in images.
These properties make SOD for spotting birds difficult.
Furthermore, various real-world applications use SOD for spotting birds, as mentioned in Sec.~\ref{section:applications}.

The contributions of this paper are as follows:
\begin{itemize}
      \item The Distant Bird Detection dataset~\cite{fujii2021distant} is extended so that more amount and variety of wild birds are observed, as shown in Fig.~\ref{fig:overview_sod4sb}.
      This extended dataset is called the Small Object Detection for Spotting Birds (SOD4SB) dataset.
      The baseline code is also provided with the dataset.
      \item The challenge-winning methods (five methods) are briefly introduced.
\end{itemize}


\section{Why Wild Birds? Sample Applications of Small Bird Detection}
\label{section:applications}

Among all possible targets in SOD, this challenge focuses on wild birds.
The application of the technology for recognizing birds in images is expected to be in the field of nature conservation and in bird damage prevention technology.

In the field of nature conservation, it is important to understand the status of bird habitats, but it has been necessary to conduct periodic surveys with the naked eye, which requires a great deal of labor. Ogawa et al. have developed a technology that automates the previously manual survey of bird populations through image recognition~\cite{DBLP:conf/igarss/OgawaLTHKM21}. Such technology will significantly reduce labor and realize efficient nature conservation activities. We look forward to further technological development by applying and developing the recognition technology tested in this competition.

Next is the application to bird damage prevention technology.
Damage caused by birds is not limited to primary industries such as agriculture~\cite{dehaven1981estimating} and fisheries~\cite{spanier1980use} but also covers a wide variety of fields such as aircraft~\cite{hedayati2015bird} and electric utility industry~\cite{huppop2006bird}, and the amount of damage is enormous.
As technology for avoiding damage, techniques that use sound and light to drive away birds are widely used, and in recent years, UAVs have been developed to drive away birds~\cite{grimm2012autonomous}.
However, conventional birds control technologies are installed continuously over a certain period of time, regardless of whether birds are present or not. As a result, birds become accustomed to them, and their effectiveness is reduced or nonexistent~\cite{mahesh2017distress}. Therefore, there is a need to develop a technology to control birds only when they are detected and to suppress the occurrence of habituation, but a technology that can recognize a wide range of field and minute birds has not been put to practical use.
By combining technologies that detect the birds with birds control technologies, it is expected that technologies that can avoid the habituation of birds will be developed, thereby reducing bird damage in a wide range of fields.


\begin{figure*}[t]
  \begin{center}
    \includegraphics[width=1\linewidth]{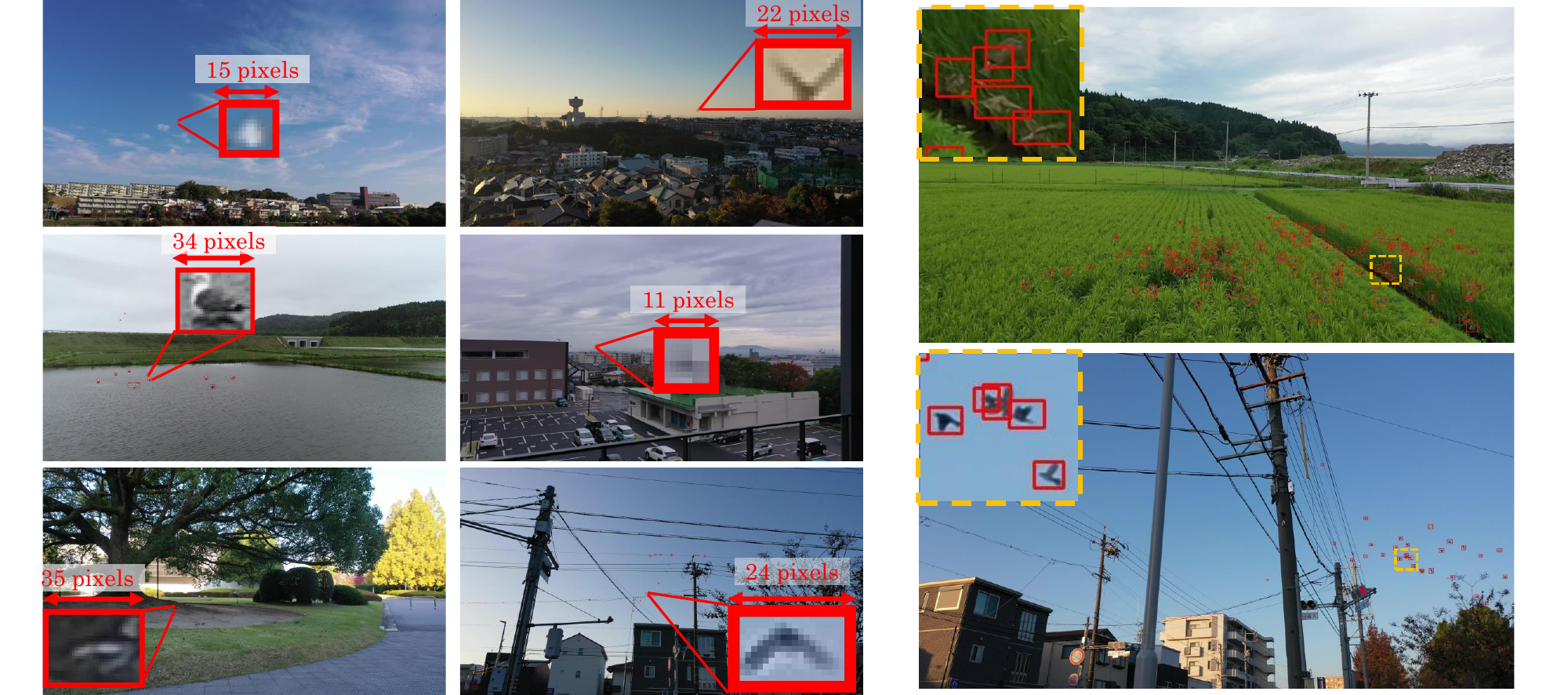}
  ~\hspace*{5mm}
    (a) Various types of small birds and diverse scenes\label{fig:small_birds} \hspace*{10mm} (b) Crowded and mutually occluded birds\label{fig:birds_crowd}  \hspace*{0mm}
    \end{center}
  \vspace{-5mm}
  \caption{Samples of SOD4SB. As shown in (a) and (b), the birds in this dataset are not only small, but also require recognition in a variety of scenes and furthermore occlude each other, making it a challenging dataset for the SOD task.}
  \label{fig:sod4sb_sample}
\end{figure*}

\section{SOD4SB Dataset}

One of the difficulties in developing SOD datasets for spotting birds is annotation cost.
Even for human annotators, it is difficult to correctly annotate small birds flying against a background of highly-textured objects such as the leaves of trees. In the Drone vs. Bird Detection Challenge~\cite{coluccia2021drone}, although drones are annotated, wild birds are not, and A. Coluccia et al. believe that the annotation of such birds is an important future challenge.
Furthermore, annotating all crowded birds is more erroneous and time-consuming.

While a few SOD datasets for spotting birds~\cite{yoshihashi2017bird,yoshihashi2015construction,sun2022airbirds} are developed, these datasets have some limitations.
In the Wind Farm dataset~\cite{yoshihashi2017bird,yoshihashi2015construction}, time-lapse images were captured from a limited number of fixed-view points.
In~the AirBird dataset~\cite{sun2022airbirds}, time-lapse images were captured from fixed-view cameras only around airports.

Our SOD4SB dataset, on the other hand, has a variety of images that are useful for various types of real-world applications, which are introduced in Sec.~\ref{section:applications}.

\subsection{Collection} 

On-drone cameras were used for image collection. The drones used for filming were the DJI Mavic 2 Pro and the DJI Phantom 4 Pro V2.0.
The camera captures videos at 30 fps, while temporal frames in the same video are regarded as independent frames in this year's challenge.
The image resolution is 3,840$\times$2,160 pixels.
The videos were captured in various locations such as urban areas, parks, forests, and fields under different weather conditions, as shown in Fig.~\ref{fig:sod4sb_sample} (a).
Birds observed in the images are hawks, crows, waterfowls, sparrows, and so on.
Several types of birds are crowded and mutually occluded, as shown in Fig.~\ref{fig:sod4sb_sample} (b).
Due to the quick motion of the birds and drone, bird and background images are sometimes blurred, as shown in Fig.~\ref{fig:overview_sod4sb}.
Since most birds were located far from the drone, most bird instances in the images are considered to be small objects, as described in Sec.~\ref{subsub:appropriateness4sod}.

\subsection{Annotation}

We manually extracted temporal frames in which any birds are observed in each video.
The extracted temporal frames were annotated so that trained annotators enclosed each bird instance by a bounding-box by the publicly-available video annotation tool, VATIC~\cite{vondrick2013efficiently}.
The annotated bounding boxes were double-checked. 
While several types of wild birds are observed in the SOD4SB dataset, all types of birds are annotated as ``bird'' because it is difficult to correctly classify all small bird instances even by human annotators.
In total, the SOD4SB dataset includes 39,070 images and 137,121 bird instances.

\subsection{Splitting}

The 39,070 annotated images and instances are split as follows:
\begin{itemize}
      \item \textbf{Training subset} consists of 9,759 images with 29,037 annotated bird instances.  
      \item \textbf{Public Test subset} consists of 9,699 images with 29,775 annotated bird instances.  
      \item \textbf{Private Test subset} consists of 20,512 images with 78,309 annotated bird instances. 
\end{itemize}
Temporal frames within the same video are considered independent for this challenge, so the images are shuffled.

The annotation data of the public test subset and both the images and annotation data of the private test subset are not publicly available.

\subsection{Appropriateness for SOD}\label{subsub:appropriateness4sod}

\begin{figure*}[t]
    \centering
    \small
    \begin{subfigure}[b]{0.329\textwidth}
        \centering
        \includegraphics[width=\textwidth]{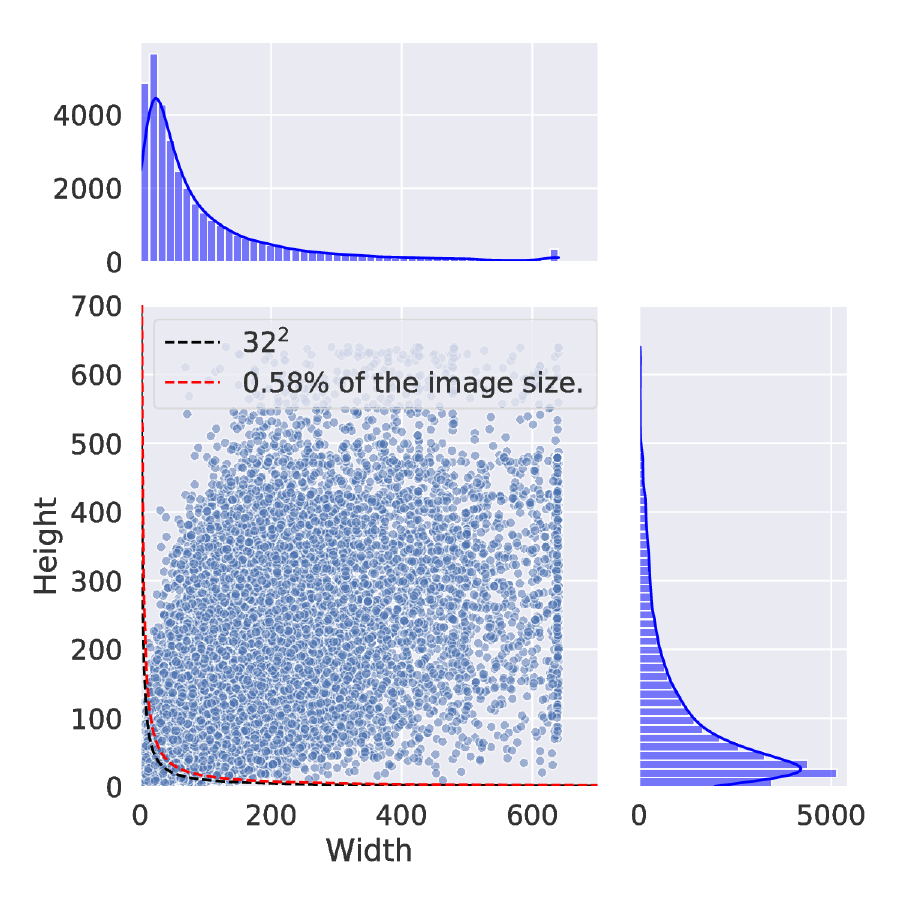}
        \caption{COCO (val2017)~\cite{COCO_ECCV2014}}
        \label{fig:coco_distribution}
    \end{subfigure}
    \hfill
    \begin{subfigure}[b]{0.329\textwidth}
        \centering
        \includegraphics[width=\textwidth]{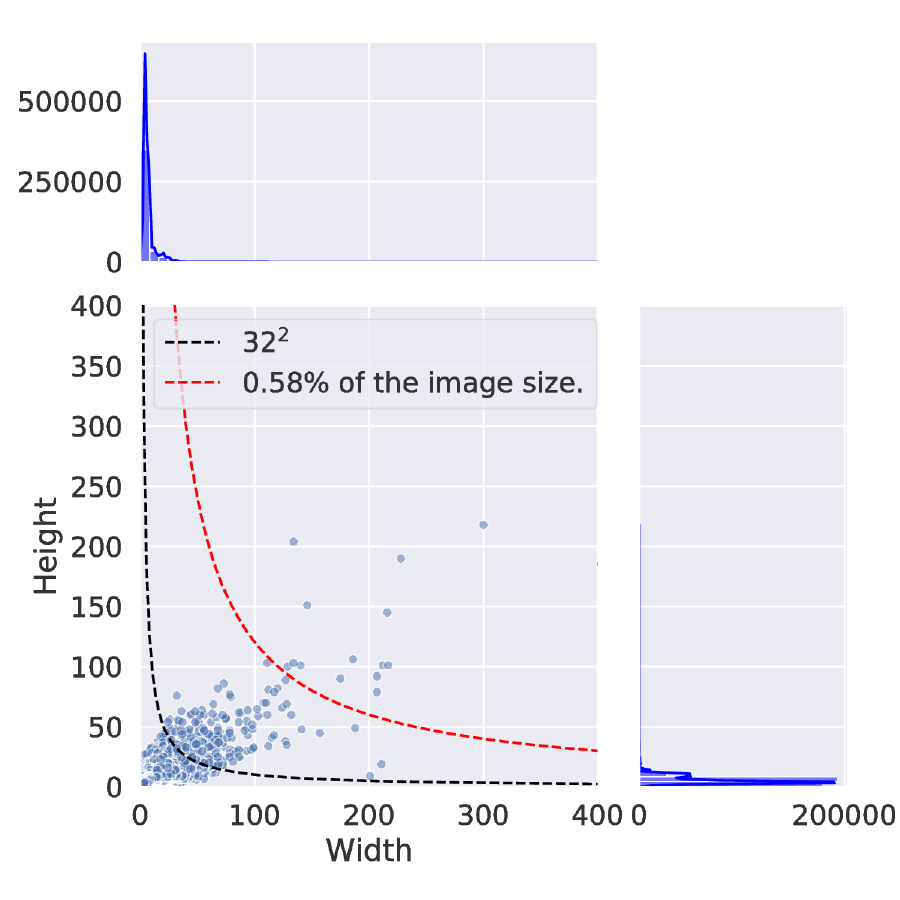}
        \caption{AirBirds~\cite{sun2022airbirds}}
        \label{fig:air_birds_distribution}
    \end{subfigure}
    \hfill
    \begin{subfigure}[b]{0.329\textwidth}
        \centering
        \includegraphics[width=\textwidth]{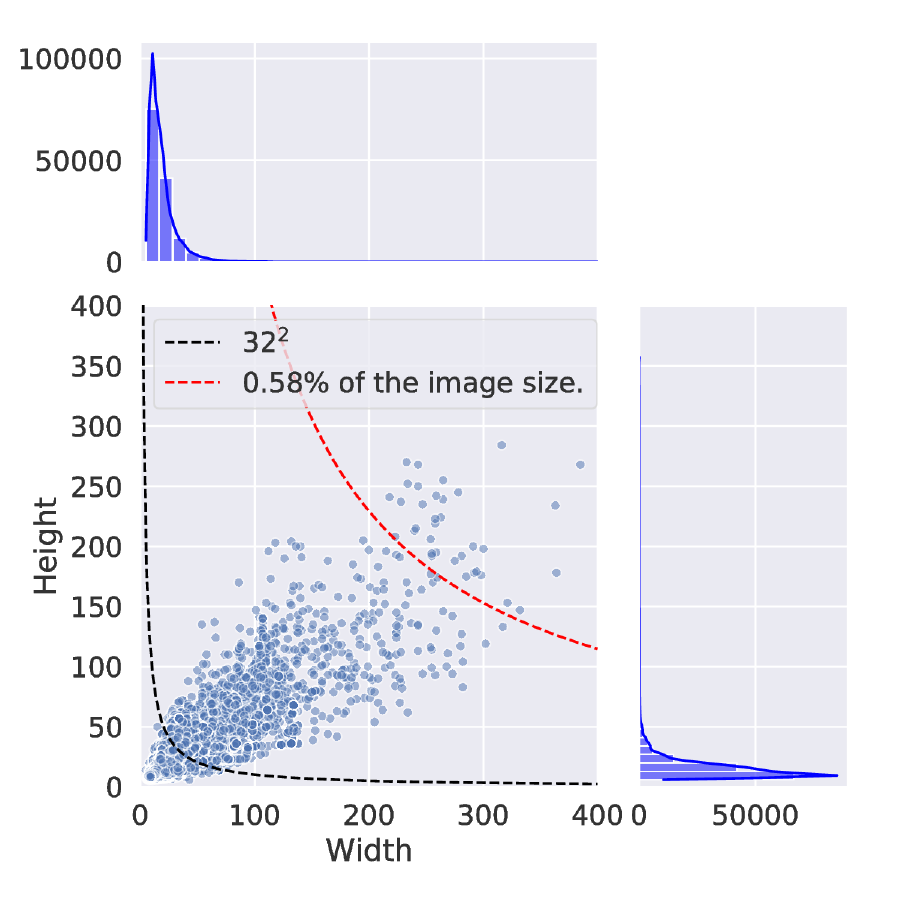}
        \caption{SOD4SB (Ours)}
        \label{fig:sod4sb_distribution}
    \end{subfigure}
    \caption{Comparison of object size distributions for (a) a generic object detection dataset, (b) an existing SOD dataset, and (c) the SOD4SB dataset. The dotted line is the definition of small object, and the image sizes used in the relative definitions are (a) 574$\times$484 pixels, (b) 1,920$\times$1,080 pixels, and (c) 3,840$\times$2,160 pixels, respectively.}
    \label{fig:dataset_comparison}
\end{figure*}

The quantitative validation is described in what follows.
While there are several criteria of SOD, our SOD4SB dataset is evaluated with two criteria.
(i)~In~\cite{torralba200880, zhu2016traffic}, the pixel size of each instance is simply evaluated so that the instance is regarded as a small object if its size is less than $32^{2}\;$pixels.
(ii)~In~\cite{chen2017r}, on the other hand, the criterion is defined with the relative sizes of objects compared to the image size so that the median of the relative sizes of all instances in each object category is less than 0.58\% of the image size.

Adapting criterion (i) to the SOD4SB dataset, the number of instances of the corresponding small objects was 74,612, meaning 95.28\% of the total.
Adapting criterion (ii) to the SOD4SB dataset, the median object size relative to the image size was $0.002\%<0.58\%$, meeting the requirement. Furthermore, the number of objects satisfying the relative size of $0.58\%$ was 78,222, meaning 99.89\%. Based on the above, we consider the SOD4SB dataset the specialized dataset for SOD, since objects that satisfy the definition of small objects are dominant in SOD4SB. 

For intuitively validating the appropriateness of the SOD4SB dataset for SOD, the distributions of the object instance sizes in the SOD4SB dataset, the MS COCO validation dataset (val2017)~\cite{COCO_ECCV2014} for generic object detection, and the AirBirds dataset~\cite{sun2022airbirds} for SOD are shown in Fig.~\ref{fig:dataset_comparison} along with the aforementioned small objects definition. Comparing the SOD4SB dataset with the COCO dataset, we can see that more objects clearly meet the definition of small objects. On the other hand, a comparison with the AirBirds dataset shows that the distribution seems to be more concentrated on smaller objects in the AirBirds dataset. However, when compared by criterion (ii) of small objects, the number of objects in the SOD does not change significantly. This is due to the different resolutions of the AirBirds and SOD4SB datasets. From this, the SOD4SB dataset is dominated by objects that satisfy the definition of small objects, and the distribution of object sizes is more diverse than AirBirds. When evaluating models on metrics such as AP, the SOD4SB dataset, which meets the definition of small objects but has a broad distribution, can appropriately evaluate models that can detect small objects of various sizes rather than overestimating models that can only detect extremely small objects.

\section{MVA2023 Small Object Detection Challenge for Spotting Birds}

We describe the details of the challenge using the SOD4SB dataset planned by the organizers.

\subsection{Baseline Code}

\begin{table}[t]
  \centering
  \caption{Comparison of difficulty levels for datasets based on CenterNet~\cite{zhou2019objects} with the ResNet18~\cite{he2016deep}-based backbone. COCO (val2017) scores taken from~\cite{zhou2019objects}.}
    \small
    \begin{tabular}{l|rrr}
    \toprule
    Dataset & \multicolumn{1}{c}{\textbf{AP@50}} & \multicolumn{1}{c}{AP@25} & \multicolumn{1}{c}{AP@75} \\
    \midrule
    \midrule
    SOD4SB public test & \textbf{46.4} & 59.5  & 5.4 \\
    SOD4SB private test & \textbf{15.4} & 24.1  & 1.6 \\
    \midrule
    COCO (val2017)~\cite{COCO_ECCV2014} & \textbf{51.5} & - & 35.1\\
    \bottomrule
    \end{tabular}%
  \label{tab:baseline_code}%
\end{table}%

The organizers provide the baseline code~\cite{baselinecode_mva2023_sod_challenge} for the challenge.
This code is developed based on CenterNet~\cite{zhou2019objects} with the ResNet18~\cite{he2016deep}-based backbone provided in MMDetection~\cite{mmdetection}.
The network is trained with hard negative mining to cope with an imbalance problem in which foreground pixels are significantly less than background pixels.

The results of bird detection using the baseline code are shown in Table~\ref{tab:baseline_code}.
The detection performance on the SOD4SB public test and private test are much worse than Centernet's AP@50 on the MS COCO validation (val2017)~\cite{COCO_ECCV2014}: 46.4 and 15.4 vs. 51.5~\cite{zhou2019objects}. The difference at AP@75 is even more pronounced.
This comparison proves the difficulty in SOD on our SOD4SB dataset and this challenge.

\subsection{Challenge Phases}

The public and private test phases were given to participants.
In the public test phase, the participants can evaluate their methods on the public test subset of the SOD4SB dataset by submitting the detection result to CodaLab~\cite{codalab_competitions}.
In this phase, the participants can access to only images without annotations.
In the private test phase, the organizer ran the code provided by each team for evaluation on the private test subset of the SOD4SB dataset.
After the challenge ends also, CodaLab is publicly available for evaluation on the public test subset, as described in Note~\ref{footnote:codalab} in the footnote.
Each team can evaluate their results at most two times per day for restricting HARKing~\cite{kerr1998harking} to the public test subset.

\subsection{Challenge Categories and Ranking Criteria}
\label{subsection:categories}

Our challenge has two categories.
In the development category, participants are requested to improve the AP@50 score on the private test subset.
Only the score is evaluated.
No technical novelty is appraised for the ranking in the development category.
In addition to the AP@50 score, technical novelties and methodological effectiveness are evaluated in the research category.
This evaluation is done by three reviewers per paper in a double-blind manner and given an average score ranging from 0 to 5 points.


\section{Challenge Results}

\begin{table*}[t!]
  \centering
  \caption{Quantitative evaluation results from public and private tests for this challenge. In the category column, ``R'' represents the research category and ``D'' represents the development category. Runtime indicates the inference time for one image when the mini-batch size is set to 1. The results of the public test are the final results after the challenge period.}
    \resizebox{\linewidth}{!}{
    \begin{tabular}{c|c|l|rrr|rrr|c|r|r|l}
    \toprule
    \multirow{2}[2]{*}{Category} &     \multirow{2}[2]{*}{Rank} & \multicolumn{1}{c|}{\multirow{2}[2]{*}{Team}} & \multicolumn{3}{c|}{Private Test} & \multicolumn{3}{c|}{Public Test} & \multicolumn{1}{c|}{\multirow{2}[2]{*}{\textbf{Review}}} & \multicolumn{1}{c|}{\multirow{1}[1]{*}{Params.}} & \multicolumn{1}{c|}{Runtime} & \multicolumn{1}{c}{\multirow{2}[2]{*}{GPU}} \\
     &     &       & \multicolumn{1}{c}{\textbf{AP@50}} & \multicolumn{1}{c}{AP@25} & \multicolumn{1}{c|}{AP@75} & \multicolumn{1}{c}{AP@50} & \multicolumn{1}{c}{AP@25} & \multicolumn{1}{c|}{AP@75} &       & \multicolumn{1}{c|}{[M]} & \multicolumn{1}{c|}{[s/image]} &  \\
    \midrule
    \midrule
    \multirow{5}[2]{*}{R} & 1     & Elsa Lab Team & \textbf{30.3} & 42.9  & 7.5   & 77.6  & 84.0  & 22.5  & \multicolumn{1}{r|}{\textbf{4.33}} & 1600  & 77.00 & V100 \\
          & 2     & Happy Day & \textbf{22.6} & 35.8  & 7.2   & 70.2  & 80.5  & 14.0  & \multicolumn{1}{r|}{\textbf{4.67}} & 70    & 0.26  & A100\\
          & 3     & Yosuke Shinya & \textbf{23.7} & 36.0  & 5.6   & 73.1  & 80.2  & 19.1  & \multicolumn{1}{c|}{\textbf{Secret}} & 32    & 0.27  & RTX3090 \\
          & 4     & DL team & \textbf{22.9} & 31.8  & 5.3   & 73.1  & 80.2  & 19.1  & \multicolumn{1}{r|}{\textbf{2.67}} & 26    & 0.09  & RTX3090 \\
          & 5     & Syusuke Yasui (E2) & \textbf{22.1} & 30.8  & 5.2   & 69.6  & 78.1  & 19.2  & \multicolumn{1}{c|}{\textbf{Secret}} & 32    & 6.80  & A100 \\
    \midrule
    \multirow{5}[2]{*}{D} & 1     & sgm   & \textbf{27.2} & 35.8  & 7.2   & 73.7  & 80.3  & 20.4  & -     & 36    & 1.00  & A30 \\
          & 2     & rakhaja & \textbf{24.7} & 32.1  & 5.6   & 69.3  & 74.3  & 18.3  & -     & 31    & 4.39  & RTX5000 \\
          & 3     & k-takasan & \textbf{23.7} & 34.8  & 2.3   & 70.7  & 80.8  & 12.1  & -     & 150   & 2.00  & V100 \\
          & 4     & JaperNing & \textbf{19.9} & 31.8  & 2.8   & 59.5  & 71.7  & 10.1  & -     & 130   & 0.70  &  RTX2000\\
          & 5     & GloryRoad & \textbf{19.1} & 28.2  & 2.7   & 67.9  & 78.5  & 13.8  & -     & 100   & 4.00  & A100 \\
    \bottomrule
    \end{tabular}%
    }
  \label{tab:result_overall}%
\end{table*}%

223 participants joined this challenge.
Based on the ranking criteria described in Sec.~\ref{subsection:categories}, five teams are selected as winners in both the development and research categories, as shown in Table~\ref{tab:result_overall}.
While the results of many teams are above those of the baseline code in AP@50, the results on the private test set are not as different as the public test compared to the baseline code scores and are significantly down from the public test scores for each method.
This is presumably due to a gap between the data distribution of the public and private test subsets, but since both images were taken during the same time period and at the same location, this difference should be minute.
Even under these conditions, since the AP@50 score of the Elsa Lab Team which won the first rank in the research category is above 30, their approach can be considered a method with high generalization capacity.

\begin{figure}[t]
  \centering
  \includegraphics[width=1.05\linewidth]{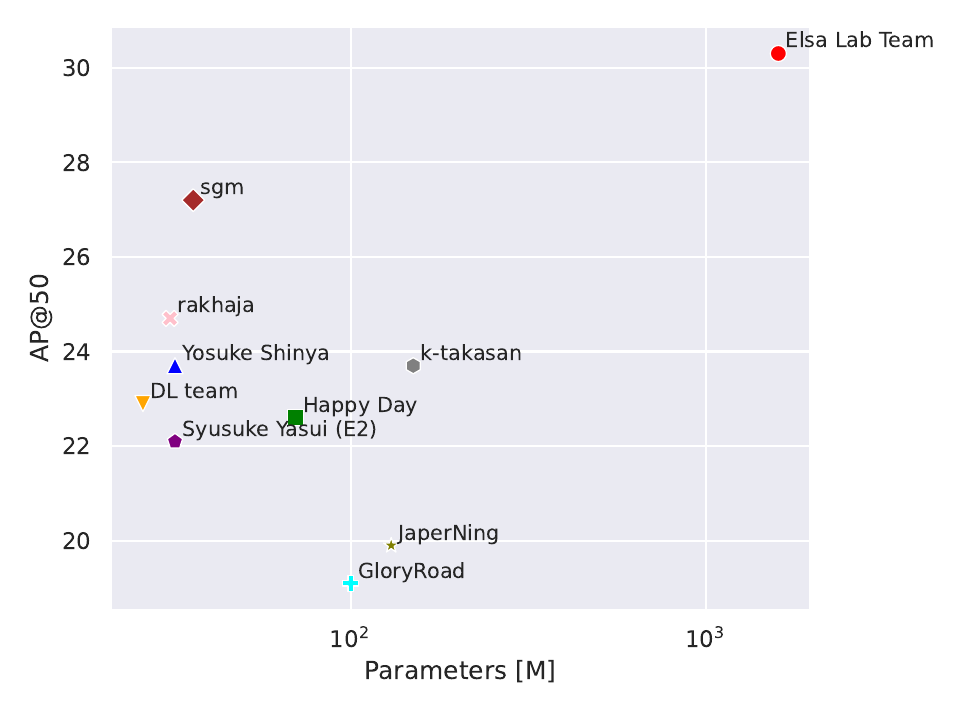}
  \vspace{-7mm}
  \caption{
    Parameters-AP@50 trade-off of winner's solutions in SOD4SB private test subset. The horizontal axis is shown in log scale.
  }
  \vspace{-0.8em}
  \label{fig:params_score}
\end{figure}

Figure~\ref{fig:params_score} shows the relationship between the AP@50 score and model size of the challenge-winning 10 methods. The Elsa Lab Team achieved the highest AP with a huge number of parameters, while sgm combines an efficient model with a low number of parameters and a high AP.

\section{Challenge Methods and Teams}

This section briefly describes the methods proposed by the winning teams in Research Category. 


\begin{figure*}[t]
    \centering
    \small
    \begin{subfigure}[b]{0.622\textwidth}
        \centering
        \includegraphics[width=\textwidth]{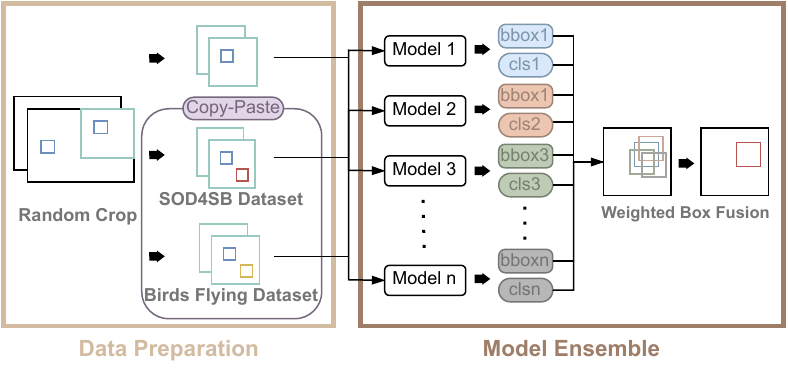}
        \caption{Overview of the framework}
        \label{fig:workflow_team1}
    \end{subfigure}
    \hfill
    \begin{subfigure}[b]{0.35\textwidth}
        \centering
        \includegraphics[width=\textwidth]{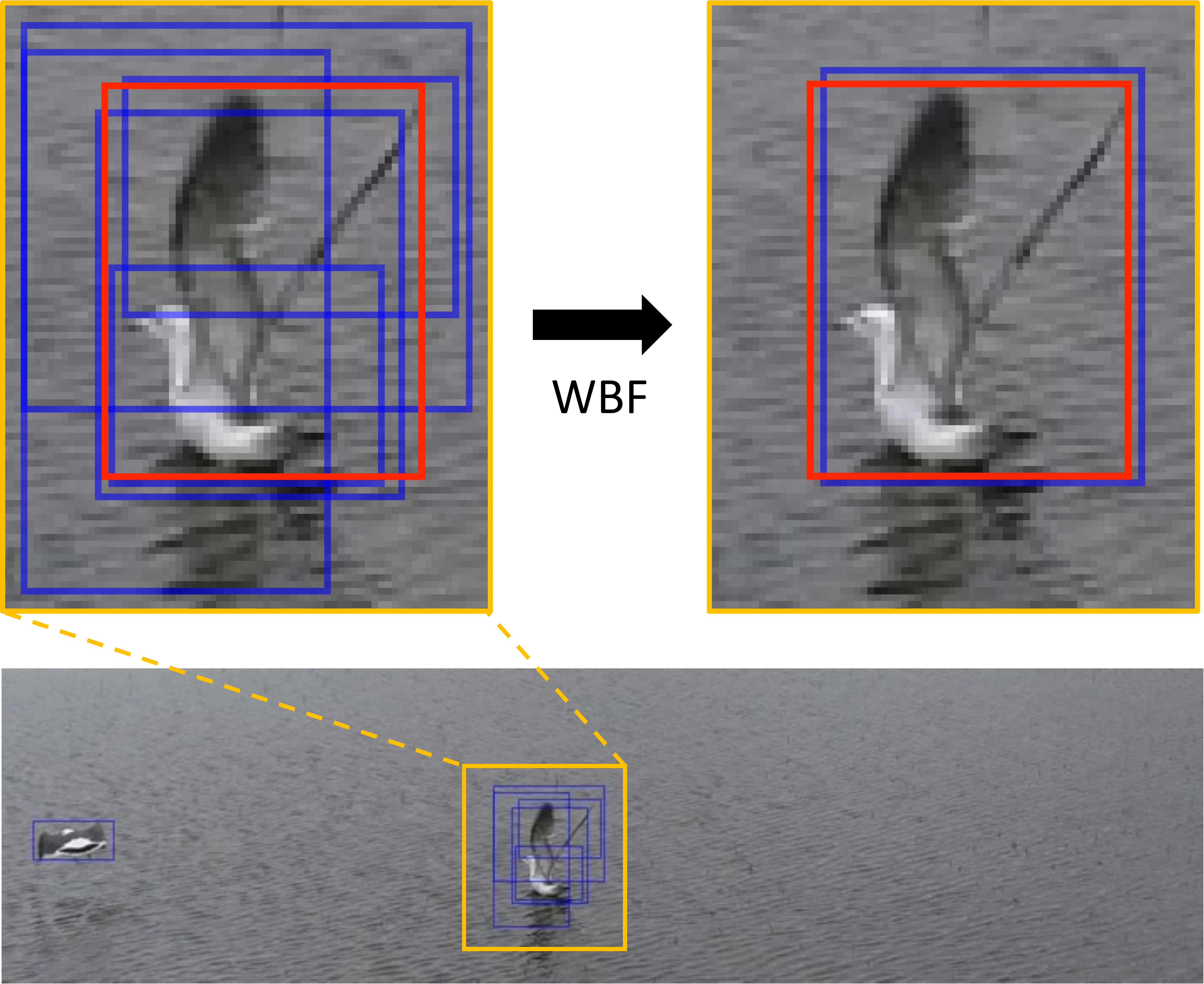}
        \caption{Visualization of the impact of WBF}
        \label{fig:wbf_effect_team1}
    \end{subfigure}
    \caption{(a) An overview of the framework~\cite{team1_mva2023}; (b) Visualization of the impact of WBF: 
    Comparison of the predictions before and after applying WBF (indicated by the \textcolor{blue}{blue} boxes) against the ground truth (depicted by the \textcolor{red}{red} box).}
    \label{fig:overview_team1}
\end{figure*}

\subsection{Elsa Lab Team (Team1)}
Elsa Lab Team (Team1) utilized an ensemble fusion method~\cite{team1_mva2023} that leverages the strengths of existing approaches to enhance the overall performance. To achieve this objective, their ensemble fusion method integrates variants from two model architectures: Cascade R-CNN~\cite{cai2018cascade} and CenterNet~\cite{zhou2019objects}.
During the training phase, an assortment of backbones (e.g., InternImage~\cite{wang2023internimage} and ResNet~\cite{he2016deep}) and techniques (e.g., Normalized Wasserstein Distance (NWD)~\cite{wang2021normalized} and Copy-Paste (CP)~\cite{ghiasi2021simple}), are utilized to generate variants exhibiting diverse performance attributes. In the inference phase, additional variants are produced using techniques such as Slicing Aided Hyper Inference (SAHI) \cite{akyon2022slicing} and test time augmentation (TTA). By ensembling the variants and their predictions using the Weighted Box Fusion method (WBF) \cite{solovyev2021weighted}, a substantial improvement is attained as compared to each top-performing model.

Fig.~\ref{fig:overview_team1}~(a) illustrates an overview of their proposed framework, which consists of two distinct stages: the \textit{data preparation stage} and the \textit{model ensemble stage}. In the \textit{data preparation stage}, they utilize the CP data augmentation technique to enrich the training data provided by SOD4SB. In this stage, the images from the SOD4SB dataset undergo cropping and augmentation with birds sourced from either the SOD4SB dataset or the Birds Flying dataset~\cite{birdsflyingdataset}. The augmented data are then forwarded to the \textit{model ensemble stage}, where several model variants are developed and grouped together to form an ensemble using WBF. By combining the predictions from different variants,  WBF enables the exploitation of these outputs to generate more precise final bounding boxes.

The performance of various ensembling methods and the top-performing single model 
are reported in Table~\ref{table:team1}~(b), while baseline results are shown in Table \ref{table:team1}~(a). The WBF ensembling method surpasses all baselines in terms of AP scores, achieving an AP@50 score of 77.6. Fig.~\ref{fig:overview_team1}~(b) depicts the impact of WBF, where the bounding boxes from different predictions are ensembled, resulting in a more accurate prediction.

\begin{table}[t]
    \caption{\small{The AP(\%) scores of: (a) baselines and (b) various ensemble methods evaluated on the SOD4SB testing set.}}
    \label{table:team1}
    \centering
    \small
    \begin{adjustbox}{max width=.475\textwidth}
        (a)
        \begin{tabular}{lcccc}
        \toprule
        Baseline Model      & Backbone Network    & AP@25 & \textbf{AP@50} & AP@75 \\
        \midrule
        DetectoRS~\cite{qiao2021detectors}           & ResNet-50           & 48.3    & 34.6               & 3.8    \\
        CenterNet~\cite{zhou2019objects}~\cite{baselinecode_mva2023_sod_challenge}           & ResNet-18           & 61.6    & 49.1               & 7.1    \\
        Cascade R-CNN~\cite{cai2018cascade}        & ResNet-50           & \textbf{63.1}    & \textbf{53.3}               & \textbf{10.8}    \\
        \bottomrule
    \end{tabular}
    \end{adjustbox}
    \begin{adjustbox}{max width=.475\textwidth}
        (b)
        \begin{tabular}{lccc}
        \toprule
        Ensemble Method          & AP@25 & \textbf{AP@50} & AP@75 \\
        \midrule
        \multirow{1}{5.75cm}{Top-Performing Single Model} & 80.3    & 73.7               & 18.3    \\
        Pure NMS with no weight    & 67.3    & 61.6               & 19.5    \\
        Weighted NMS               & 81.4    & 75.1               & 19.3    \\
        Soft NMS                   & 79.7    & 73.9               & 20.8    \\
        WBF (Elsa Lab Team)                        & \textbf{84.0}    & \textbf{77.6}               & \textbf{22.5}    \\
        \bottomrule
    \end{tabular}
    \end{adjustbox}
    \vspace{-1em}
\end{table}

\subsection{Happy Day}

\begin{figure*}[t]
  \begin{center}
    \includegraphics[width=1.15\linewidth,trim=90 640 0 60,clip]{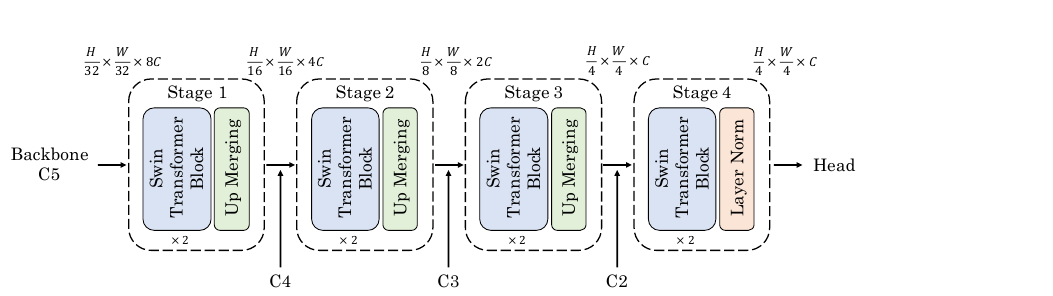}
  \end{center}
  \vspace{-1.5em}
  \caption{Happy Day team proposed neck network~\cite{happy_day_mva2023}, the windows size 2 was used in each Swin Transformer Block, and Up Merging~\cite{shi2016real} Module upsamples features and merges with those extracted in the backbone to effectively
recognize object features in SOD.}
\vspace{-0.8em}
  \label{fig:method_overview}
\end{figure*}

Happy Day team proposed a Swin Transformer~\cite{liu2021swin} based network~\cite{happy_day_mva2023} with a hierarchical design for small object detection (Fig.~\ref{fig:method_overview}), which improves the features learned by a neck network corresponding to the CenterNet~\cite{zhou2019objects} architecture to learn effective features for small objects. The key idea in Happy Day's work is to change the size of the shifting windows of the neck to a smaller one, which contributes to capture the attentions of small objects inside the small windows, which reduces the parameters and leads to good performance for small object detection. In addition, to detect small objects precisely in location even through several up-and-down-samplings, Happy Day uses skip connections~\cite{ronneberger2015u} for providing precise locations from backbone to the neck. 
The architecture of the proposed neck network is shown in Fig.~\ref{fig:method_overview}.

The input consists of different scales of features from backbone, as shown in Fig.~\ref{fig:method_overview}, from C5 (32{$\times$} down-sampling) to C2 (4{$\times$} down-sampling) with different sizes, where C$i$ ($2 \leq i \leq 5$) corresponds to $i$-th blocks of the backbone. The shifting windows are inside of the Swin Transformer blocks in each stage, and the windows size is selected as a smaller one, with a default size of 2. Further, Happy Day adds skip-connection after Stages 1, 2, and 3 for merging features with the backbone outputs. The output of the neck is passed to the CenterNet head for predicting the center points, height, and width of objects in the image.

To evaluate the effectiveness and feasibility of the proposed network, Happy Day compared the proposed neck with the original CenterNet neck, by computing the AP metrics on the validation set of the SOD4SB dataset. Experiments indicate that the most metrics for object detection, including AP@50 and AP@75, of the proposed method surpassed the ones with the default CenterNet neck.

For the key idea in Happy Day's work, the small windows size is expected to be an effective feature representation. Happy Day also performed ablation study by changing the windows sizes 2, 3, and 5 for experiments. They conducted experimental verification and found that a windows size of 2 significantly emphasized local information in comparison to other sizes, thereby leading to improved detection performance. This particular windows size is a crucial factor in efficiently computing local features, which play a vital role in small object detection.


\subsection{Yosuke Shinya}

Yosuke Shinya proposed BandASAP~\cite{BandRe_MVAW2023}, which is a set of scale-wise metrics for object detection evaluation.
For finer scale-wise evaluation than the COCO metrics~\cite{COCO_ECCV2014, cocoapi},
it is based on ASAP~\cite{USB_Shinya_BMVC2022}.
For more reliable and intuitive evaluation than ASAP,
the author proposed a filter bank consisting of triangular and trapezoidal band-pass filters.

\begin{figure}[t]
  \centering
  \includegraphics[width=\linewidth]{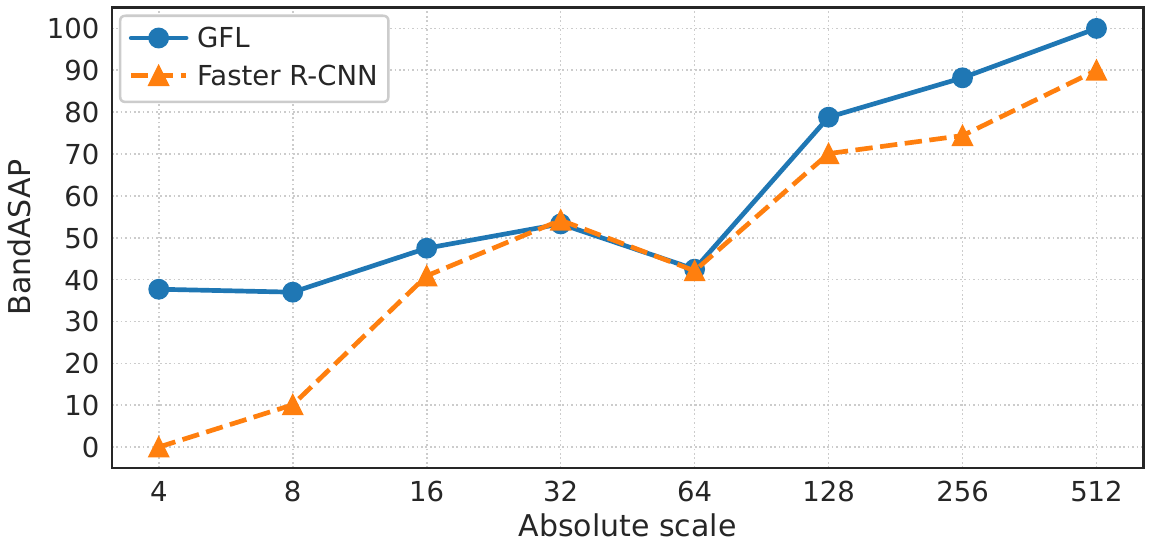}
  \caption{
    Results of BandASAP~\cite{BandRe_MVAW2023} metrics.
  }
  \label{fig:band_asap_sod4bird}
  \vspace{-1.5em}
\end{figure}

The author trained GFL~\cite{GFL_NeurIPS2020} and Faster R-CNN~\cite{Faster_R-CNN_NIPS2015} with simple settings
and selected GFL for the final submission because it is significantly better than Faster R-CNN.
He analyzed the results using the proposed metrics (Fig.~\ref{fig:band_asap_sod4bird}).
BandASAP succeeds in highlighting differences between the methods.
Although the author discussed a possible cause of low BandASAP$_{64}$, the cause of the remarkable differences between GFL and Faster R-CNN remains unknown.
Further analysis in future research would lead to performance improvement in small object detection.

\subsection{DL team}

\begin{figure*}[t]
  \begin{center}
    \includegraphics[width=0.9\linewidth]{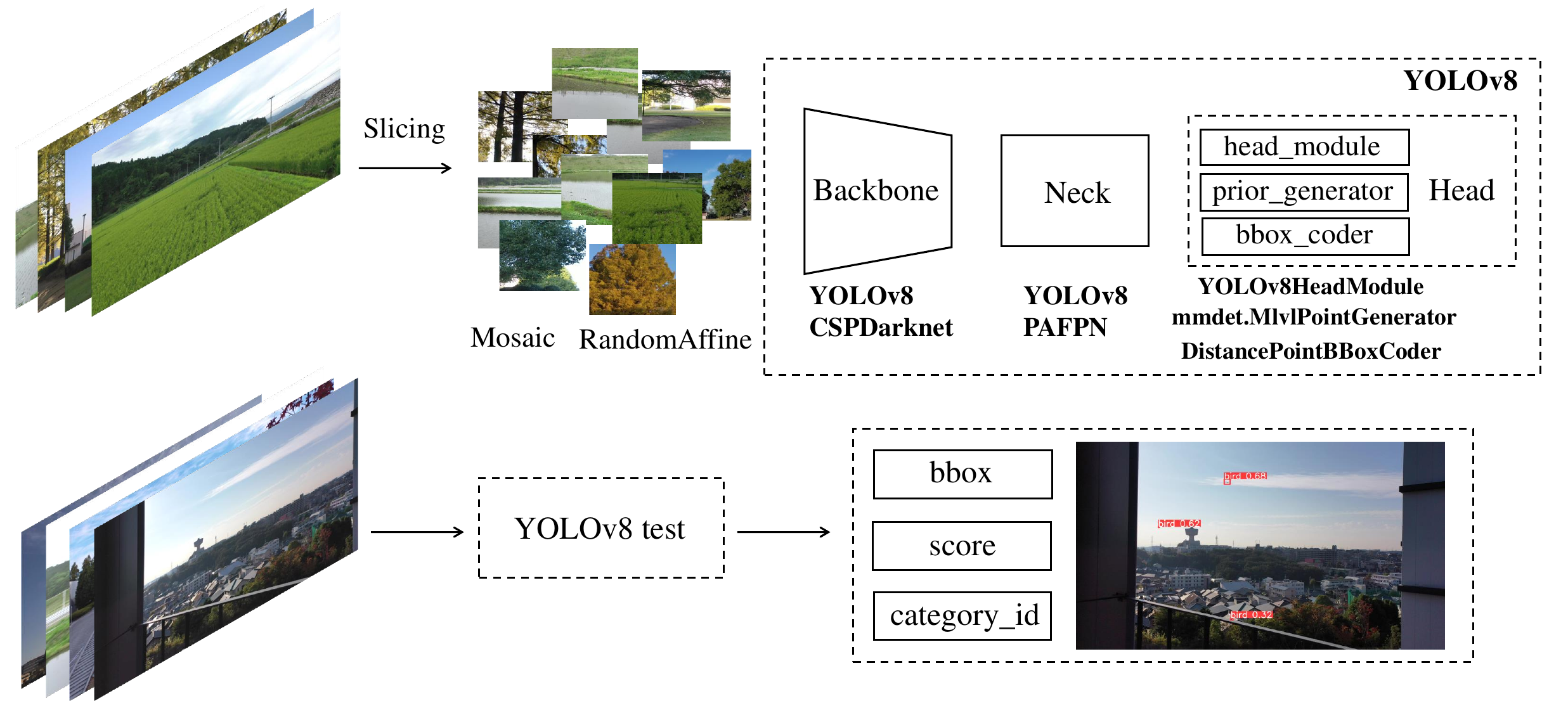}
  \end{center}
  \vspace{-1em}
  \caption{DL Method overview.}
  \label{fig:method_overview_dl}
\end{figure*}

DL team proposed a DL Method (Fig.~\ref{fig:method_overview_dl}) to enhance the detection capability of small objects. By partitioning the images in the training set into smaller sub-images for training, the method enables training with a larger batch size within the same GPU memory. This enables the model to observe a broader range of features during a single training iteration, leading to a significant improvement in the model's generalization ability and better capture of the details and features of small objects. Under comparable memory consumption conditions, the method achieves an improvement of over 7 percentage points. The partitioning method takes into account the dataset characteristics to adjust the overlap rate and annotation for the partitions. Furthermore, DL team discovers the importance of data augmentation in training small object detection networks. Consistent utilization of data augmentation in experiments enables the model to learn more features and details across diverse scenes, thereby enhancing the effectiveness of object detection. The partitioning approach generates a greater number of small object samples, while the data augmentation technique ensures sufficient diversification to simulate various variations in real-world scenarios, thus enhancing the model's robustness. Finally, DL team integrate their method into the medium-scale YOLOv8 \cite{yolo}, evaluate it on the SOD4SB public test and achieve an AP@50 score of 73.3.

\subsection{(E2) Syusuke Yasui}

(E2) Syusuke Yasui proposed 5 simple but effective methods to do small object detection.

\begin{itemize}
\item NWD (Normalized Gaussian Wasserstein Distance)~\cite{wang2021normalized,nwd2} is an improved method for objects with small IoU loss. Introduced to more accurately indicate bounding box distances for small objects. The optimal weight for loss is 3/4 of varifocal loss and 2 for refine.
\item Probability Distribution Surface Models such as CenterNet~\cite{zhou2019objects} and VarifocalNet~\cite{zhang2021varifocalnet} that predict the pixels of the object you want to detect with a probability distribution are more suitable. This is also shown by the experimental results. If the detected object is small, the ratio of positive and negative is severely unbalanced, and it is generally difficult for the model to learn the detection points.
\item Switch Hard Augmentation: By performing hard augmentation in the first half of learning and performing lightweight augmentation with only flip in the second half of learning, it is possible to create a stable, high-speed, and highly accurate model. The hard augmentation here is mosaic~\cite{wang2021scaled}, mixup~\cite{zhang2017mixup}, affine transform. Because hard augmentation effectively increases positives.
\item Multi scale train is a standard method of learning while changing the resolution with resize. Randomly selecting around 20\% of the input resolution will increase the detection accuracy the most.
\item Weight Moving Average is a method to limit the model weight by exponential moving average. The purpose of this is to avoid overfitting in a single backward step and not disproportionately overfitting, and it is effective even for small objects. Its value is 1e-4.
\end{itemize}

\section{Conclusion}
This paper proposes a new SOD dataset, the SOD4SB dataset, and reviews the MVA2023 Small Object Detection Challenge for Spotting Birds, which utilizes this dataset. The 223 participants were tasked with detecting small birds in a variety of scenes. The winners' methods performed remarkably well on the challenging SOD4SB dataset and provided several novel and progressive proposals to help solve the SOD task. We hope that the results of this challenge will help build a foundation for advanced UAVs applications.

This challenge is expected to be extended to Video SOD~\cite{rekavandi2022guide,bosquet2021correlation} or Video Small Object Tracking~\cite{zhu2023tiny,zhang2022tracking}. This extension is expected to promote research and development of more useful recognition processing for improving the accuracy of detecting small birds and for autonomous control of UAVs at a later stage. Furthermore, by setting constraints on inference time and the arithmetic unit, we expect to develop technologies that enable real-time inference on edge devices mounted on UAVs.

\section{Acknowledgments}
We would like to express our deepest gratitude to Zhao Kaikai, Riku
Miyata, and Kazutoshi Akita at Toyota Technological Institute for
their hard work in preparing for the dataset, base code, and website
for this challenge. We also would like to thank Masatsugu Kidode at
Nara Institute of Science and Technology for his helpful advice. We
also appreciate code testers and annotators.

This challenge was supported by a donation from the MVA organization.

\appendix
\section{Teams and Affiliations}
\label{sec:apd:team}

\subsection*{MVA2023 Small Object Detection Challenge for Spotting Birds Organizers}
\noindent{\textbf{Title: }}\\ MVA2023 Small Object Detection Challenge for Spotting Birds: Dataset, Methods, and Results\\
\noindent{\textbf{Members:}}\\ \textbf{Yuki Kondo$^1$ (yuki\_kondo\_ab@mail.toyota.co.jp)}, Norimichi Ukita$^2$, Takayuki Yamaguchi$^3$\\
\noindent{\textbf{Affiliations: }}\\
$^1$ Toyota Motor Corporation, Japan\\
$^2$ Toyota Technological Institute, Japan\\
$^3$ Iwate Agricultural Research Center, Japan\\

\subsection*{Elsa Lab Team (Team1)}

\noindent{\textbf{{Title:}}}\\Ensemble Fusion for Small Object Detection

\noindent{\textbf{{Members:}}}\\\textbf{Hao-Yu~Hou}\footnotemark[1]\newline(howard.hou.fan@elsa.cs.nthu.edu.tw),
{Mu-Yi~Shen}\footnotemark[1],
{Chia-Chi~Hsu}\footnotemark[1],
{En-Ming~Huang}\footnotemark[1],
{Yu-Chen~Huang}\footnotemark[1],
{Yu-Cheng~Xia}\footnotemark[1],
Chien-Yao~Wang\footnotemark[2], and
Chun-Yi~Lee\footnotemark[1].

\noindent{\textbf{{Affiliations:}}}\\
\noindent\footnotemark[1]~Elsa Lab, Department of Computer Science, National Tsing Hua University, Taiwan.\\
\noindent\footnotemark[2]~Institute of Information Science, Academia Sinica, Taiwan

\subsection*{Happy Day}
\noindent{\textbf{Title: }}\\Small Object Detection for Bird with Swin Transformer\\
\noindent{\textbf{Members:}}\\ \textbf{Da Huo$^1$ (huod@cs.is.i.nagoya-u.ac.jp)}, Marc A. Kastner$^2$, Tingwei Liu$^1$, Yasutomo Kawanishi$^{3,1}$, Takatsugu Hirayama$^{4,1}$, Takahiro Komamizu$^1$, Ichiro Ide$^1$\\
\noindent{\textbf{Affiliations: }}\\
$^1$ Nagoya University, Japan\\
$^2$ Kyoto University, Japan\\
$^3$ GRP, RIKEN, Japan\\
$^4$ University of Human Environments, Japan\\

\subsection*{Yosuke Shinya}
\noindent{\textbf{Title: }}\\ BandRe: Rethinking Band-Pass Filters for Scale-Wise Object Detection Evaluation\\
\noindent{\textbf{Members:}}\\ \textbf{Yosuke Shinya$^1$ (https://shinya7y.github.io/)}\\
\noindent{\textbf{Affiliations: }}\\
$^1$ Independent researcher, Japan\\

\subsection*{DL}
\noindent{\textbf{Title: }}\\ Method to Achieve High Performance for Small Object Detection\\
\noindent{\textbf{Members:}}\\ \textbf{Xinyao Liu$^1$ (2205124667@stu.xjtu.edu.cn)}, Guang Liang$^1$(2204313319@stu.xjtu.edu.cn)\\
\noindent{\textbf{Affiliations: }}\\
$^1$ Xi'an Jiaotong University, China

\subsection*{(E2) Syusuke Yasui}
\noindent{\textbf{Title: }}\\ More easy framework of small object detection\\
\noindent{\textbf{Members:}}\\ \textbf{Syusuke Yasui$^1$ (syuchimu@gmail.com)}\\
\noindent{\textbf{Affiliations: }}\\
$^1$ Space Shift Inc., Japan\\

\bibliographystyle{ieeetr}
\bibliography{reference.bib}

\begin{thebibliography}{10}

\bibitem{sodbchallenge2023misc}
Y.~Kondo, N.~Ukita, and T.~Yamaguchi, ``{MVA2023 Small Object Detection
  Challenge for Spotting Birds}.''
  \url{https://www.mva-org.jp/mva2023/challenge}, 2023.

\bibitem{baselinecode_mva2023_sod_challenge}
K.~Zhao, R.~Miyata, Y.~Kondo, and K.~Akita, ``{Baseline code for SOD4SB by
  IIM-TTIJ}.''
  \url{https://github.com/IIM-TTIJ/MVA2023SmallObjectDetection4SpottingBirds},
  2023.

\bibitem{codalab_competitions}
A.~Pavao, I.~Guyon, A.-C. Letournel, X.~Baró, H.~Escalante, S.~Escalera,
  T.~Thomas, and Z.~Xu, ``Codalab competitions: An open source platform to
  organize scientific challenges,'' {\em Technical report}, 2022.

\bibitem{Faster_R-CNN_NIPS2015}
S.~Ren, K.~He, R.~Girshick, and J.~Sun, ``{Faster R-CNN}: Towards real-time
  object detection with region proposal networks,'' in {\em NIPS}, 2015.

\bibitem{yolo}
G.~Jocher, A.~Chaurasia, and J.~Qiu, ``{YOLO by Ultralytics}.''
  \url{https://github.com/ultralytics/ultralytics}, 2023.

\bibitem{duan2019centernet}
K.~Duan, S.~Bai, L.~Xie, H.~Qi, Q.~Huang, and Q.~Tian, ``{CenterNet}: Keypoint
  triplets for object detection,'' in {\em ICCV}, 2019.

\bibitem{tan2020efficientdet}
M.~Tan, R.~Pang, and Q.~V. Le, ``Efficientdet: Scalable and efficient object
  detection,'' in {\em CVPR}, 2020.

\bibitem{detr}
N.~Carion, F.~Massa, G.~Synnaeve, N.~Usunier, A.~Kirillov, and S.~Zagoruyko,
  ``End-to-end object detection with transformers,'' in {\em ECCV}, 2020.

\bibitem{wang2023internimage}
W.~Wang, J.~Dai, Z.~Chen, Z.~Huang, Z.~Li, X.~Zhu, X.~Hu, T.~Lu, L.~Lu, H.~Li,
  {\em et~al.}, ``Internimage: Exploring large-scale vision foundation models
  with deformable convolutions,'' in {\em CVPR}, 2023.

\bibitem{fang2023eva}
Y.~Fang, W.~Wang, B.~Xie, Q.~Sun, L.~Wu, X.~Wang, T.~Huang, X.~Wang, and
  Y.~Cao, ``Eva: Exploring the limits of masked visual representation learning
  at scale,'' in {\em CVPR}, 2023.

\bibitem{COCO_ECCV2014}
T.-Y. Lin, M.~Maire, S.~Belongie, J.~Hays, P.~Perona, D.~Ramanan,
  P.~Doll{\'a}r, and C.~L. Zitnick, ``{Microsoft COCO}: Common objects in
  context,'' in {\em ECCV}, 2014.

\bibitem{everingham2010pascal}
M.~Everingham, L.~Van~Gool, C.~K. Williams, J.~Winn, and A.~Zisserman, ``The
  pascal visual object classes (voc) challenge,'' {\em IJCV}, vol.~88,
  pp.~303--338, 2010.

\bibitem{sod_survey1}
N.-D. Nguyen, T.~Do, T.~D. Ngo, and D.-D. Le, ``An evaluation of deep learning
  methods for small object detection,'' {\em Journal of electrical and computer
  engineering}, vol.~2020, pp.~1--18, 2020.

\bibitem{sod_survey2}
Y.~Liang, Y.~Han, and F.~Jiang, ``Deep learning-based small object detection: A
  survey,'' in {\em ICCAI}, 2022.

\bibitem{sod_survey3}
G.~Chen, H.~Wang, K.~Chen, Z.~Li, Z.~Song, Y.~Liu, W.~Chen, and A.~Knoll, ``A
  survey of the four pillars for small object detection: Multiscale
  representation, contextual information, super-resolution, and region
  proposal,'' {\em IEEE Transactions on systems, man, and cybernetics:
  systems}, vol.~52, no.~2, pp.~936--953, 2020.

\bibitem{wang2021tiny}
J.~Wang, W.~Yang, H.~Guo, R.~Zhang, and G.-S. Xia, ``Tiny object detection in
  aerial images,'' in {\em 2020 25th International Conference on Pattern
  Recognition (ICPR)}, pp.~3791--3798, IEEE, 2021.

\bibitem{yu2020scale}
X.~Yu, Y.~Gong, N.~Jiang, Q.~Ye, and Z.~Han, ``Scale match for tiny person
  detection,'' in {\em WACV}, 2020.

\bibitem{behrendt2017deep}
K.~Behrendt, L.~Novak, and R.~Botros, ``A deep learning approach to traffic
  lights: Detection, tracking, and classification,'' in {\em ICRA}, 2017.

\bibitem{waqas2019isaid}
S.~Waqas~Zamir, A.~Arora, A.~Gupta, S.~Khan, G.~Sun, F.~Shahbaz~Khan, F.~Zhu,
  L.~Shao, G.-S. Xia, and X.~Bai, ``isaid: A large-scale dataset for instance
  segmentation in aerial images,'' in {\em CVPRW}, 2019.

\bibitem{bosquet2018stdnet}
B.~Bosquet, M.~Mucientes, and V.~M. Brea, ``Stdnet: A convnet for small target
  detection,'' in {\em BMVC}, 2018.

\bibitem{yu20201st}
X.~Yu, Z.~Han, Y.~Gong, N.~Jan, J.~Zhao, Q.~Ye, J.~Chen, Y.~Feng, B.~Zhang,
  X.~Wang, {\em et~al.}, ``The 1st tiny object detection challenge: Methods and
  results,'' in {\em ECCVW}, Springer, 2020.

\bibitem{coluccia2021drone}
A.~Coluccia, A.~Fascista, A.~Schumann, L.~Sommer, A.~Dimou, D.~Zarpalas, F.~C.
  Akyon, O.~Eryuksel, K.~A. Ozfuttu, S.~O. Altinuc, {\em et~al.},
  ``Drone-vs-bird detection challenge at ieee avss2021,'' in {\em AVSS}, IEEE,
  2021.

\bibitem{zhang2019widerperson}
S.~Zhang, Y.~Xie, J.~Wan, H.~Xia, S.~Z. Li, and G.~Guo, ``Widerperson: A
  diverse dataset for dense pedestrian detection in the wild,'' {\em IEEE
  Transactions on Multimedia}, 2019.

\bibitem{fujii2021distant}
S.~Fujii, K.~Akita, and N.~Ukita, ``Distant bird detection for safe drone
  flight and its dataset,'' in {\em MVA}, 2021.

\bibitem{DBLP:conf/igarss/OgawaLTHKM21}
K.~Ogawa, Y.~Lin, H.~Takeda, K.~Hashimoto, Y.~Konno, and K.~Mori, ``Automated
  counting wild birds on {UAV} image using deep learning,'' in {\em
  International Geoscience and Remote Sensing Symposium, {IGARSS}}, 2021.

\bibitem{dehaven1981estimating}
R.~DeHaven and R.~Hothem, ``Estimating bird damage from damage incidence in
  wine grape vineyards,'' {\em American journal of enology and viticulture},
  vol.~32, no.~1, pp.~1--4, 1981.

\bibitem{spanier1980use}
E.~Spanier, ``The use of distress calls to repel night herons (nycticorax
  nycticorax) from fish ponds,'' {\em Journal of Applied Ecology},
  pp.~287--294, 1980.

\bibitem{hedayati2015bird}
R.~Hedayati and M.~Sadighi, {\em Bird strike: an experimental, theoretical and
  numerical investigation}.
\newblock Woodhead Publishing, 2015.

\bibitem{huppop2006bird}
O.~H{\"u}ppop, J.~Dierschke, K.-M. EXO, E.~Fredrich, and R.~Hill, ``Bird
  migration studies and potential collision risk with offshore wind turbines,''
  {\em Ibis}, vol.~148, pp.~90--109, 2006.

\bibitem{grimm2012autonomous}
B.~A. Grimm, B.~A. Lahneman, P.~B. Cathcart, R.~C. Elgin, G.~L. Meshnik, and
  J.~P. Parmigiani, ``Autonomous unmanned aerial vehicle system for controlling
  pest bird population in vineyards,'' in {\em ASME International Mechanical
  Engineering Congress and Exposition}, 2012.

\bibitem{mahesh2017distress}
S.~Mahesh, V.~Vasudeva~Rao, G.~Surender, D.~Kiran~kumar, and K.~Swamy,
  ``Distress feeding of depredatory birds in sunflower and sorghum protected by
  bioacoustics,'' {\em bioRxiv}, p.~200097, 2017.

\bibitem{yoshihashi2017bird}
R.~Yoshihashi, R.~Kawakami, M.~Iida, and T.~Naemura, ``Bird detection and
  species classification with time-lapse images around a wind farm: Dataset
  construction and evaluation,'' {\em Wind Energy}, vol.~20, no.~12,
  pp.~1983--1995, 2017.

\bibitem{yoshihashi2015construction}
R.~Yoshihashi, R.~Kawakami, M.~Iida, and T.~Naemura, ``Construction of a bird
  image dataset for ecological investigations,'' in {\em ICIP}, 2015.

\bibitem{sun2022airbirds}
H.~Sun, Y.~Wang, X.~Cai, P.~Wang, Z.~Huang, D.~Li, Y.~Shao, and S.~Wang,
  ``Airbirds: A large-scale challenging dataset for bird strike prevention in
  real-world airports,'' in {\em ACCV}, 2022.

\bibitem{vondrick2013efficiently}
C.~Vondrick, D.~Patterson, and D.~Ramanan, ``Efficiently scaling up
  crowdsourced video annotation: A set of best practices for high quality,
  economical video labeling,'' {\em IJCV}, vol.~101, pp.~184--204, 2013.

\bibitem{torralba200880}
A.~Torralba, R.~Fergus, and W.~T. Freeman, ``80 million tiny images: A large
  data set for nonparametric object and scene recognition,'' {\em PAMI},
  vol.~30, no.~11, pp.~1958--1970, 2008.

\bibitem{zhu2016traffic}
Z.~Zhu, D.~Liang, S.~Zhang, X.~Huang, B.~Li, and S.~Hu, ``Traffic-sign
  detection and classification in the wild,'' in {\em CVPR}, 2016.

\bibitem{chen2017r}
C.~Chen, M.-Y. Liu, O.~Tuzel, and J.~Xiao, ``R-cnn for small object
  detection,'' in {\em ACCV}, 2017.

\bibitem{zhou2019objects}
X.~Zhou, D.~Wang, and P.~Kr{\"a}henb{\"u}hl, ``Objects as points,'' {\em
  arXiv}, 2019.

\bibitem{he2016deep}
K.~He, X.~Zhang, S.~Ren, and J.~Sun, ``Deep residual learning for image
  recognition,'' in {\em CVPR}, 2016.

\bibitem{mmdetection}
K.~Chen, J.~Wang, J.~Pang, Y.~Cao, Y.~Xiong, X.~Li, S.~Sun, W.~Feng, Z.~Liu,
  J.~Xu, Z.~Zhang, D.~Cheng, C.~Zhu, T.~Cheng, Q.~Zhao, B.~Li, X.~Lu, R.~Zhu,
  Y.~Wu, J.~Dai, J.~Wang, J.~Shi, W.~Ouyang, C.~C. Loy, and D.~Lin,
  ``{MMDetection}: Open mmlab detection toolbox and benchmark,'' {\em arXiv},
  2019.

\bibitem{kerr1998harking}
N.~L. Kerr, ``Harking: Hypothesizing after the results are known,'' {\em
  Personality and social psychology review}, vol.~2, no.~3, pp.~196--217, 1998.

\bibitem{team1_mva2023}
H.-Y. Hou, M.-Y. Shen, C.-C. Hsu, E.-M. Huang, Y.-C. Huang, Y.-C. Xia, C.-Y.
  Wang, and C.-Y. Lee, ``Ensemble fusion for small object detection,'' in {\em
  MVA}, 2023.

\bibitem{cai2018cascade}
Z.~Cai and N.~Vasconcelos, ``Cascade {R-CNN}: Delving into high quality object
  detection,'' in {\em CVPR}, 2018.

\bibitem{wang2021normalized}
J.~Wang, C.~Xu, W.~Yang, and L.~Yu, ``A normalized gaussian wasserstein
  distance for tiny object detection,'' {\em arXiv}, 2021.

\bibitem{ghiasi2021simple}
G.~Ghiasi, Y.~Cui, A.~Srinivas, R.~Qian, T.-Y. Lin, E.~D. Cubuk, Q.~V. Le, and
  B.~Zoph, ``Simple copy-paste is a strong data augmentation method for
  instance segmentation,'' in {\em CVPR}, 2021.

\bibitem{akyon2022slicing}
F.~C. Akyon, S.~Onur~Altinuc, and A.~Temizel, ``Slicing aided hyper inference
  and fine-tuning for small object detection,'' in {\em ICIP}, 2022.

\bibitem{solovyev2021weighted}
R.~Solovyev, W.~Wang, and T.~Gabruseva, ``Weighted boxes fusion: Ensembling
  boxes from different object detection models,'' {\em Image and Vision
  Computing}, pp.~1--6, 2021.

\bibitem{birdsflyingdataset}
Gareth, ``Birds flying dataset.''
  \url{www.kaggle.com/datasets/nelyg8002000/birds-flying}, 2021.

\bibitem{qiao2021detectors}
S.~Qiao, L.-C. Chen, and A.~Yuille, ``{DetectoRS}: Detecting objects with
  recursive feature pyramid and switchable atrous convolution,'' in {\em CVPR},
  2021.

\bibitem{happy_day_mva2023}
D.~Huo, M.~A. Kastner, T.~Liu, Y.~Kawanishi, T.~Hirayama, T.~Komamizu, and
  I.~Ide, ``Small object detection for bird with swin transformer,'' in {\em
  MVA}, 2023.

\bibitem{shi2016real}
W.~Shi, J.~Caballero, F.~Husz{\'a}r, J.~Totz, A.~P. Aitken, R.~Bishop,
  D.~Rueckert, and Z.~Wang, ``Real-time single image and video super-resolution
  using an efficient sub-pixel convolutional neural network,'' in {\em CVPR},
  2016.

\bibitem{liu2021swin}
Z.~Liu, Y.~Lin, Y.~Cao, H.~Hu, Y.~Wei, Z.~Zhang, S.~Lin, and B.~Guo, ``Swin
  transformer: Hierarchical vision transformer using shifted windows,'' in {\em
  ICCV}, 2021.

\bibitem{ronneberger2015u}
O.~Ronneberger, P.~Fischer, and T.~Brox, ``U-net: Convolutional networks for
  biomedical image segmentation,'' in {\em MICCAI}, 2015.

\bibitem{BandRe_MVAW2023}
Y.~Shinya, ``{BandRe}: Rethinking band-pass filters for scale-wise object
  detection evaluation,'' in {\em MVA}, 2023.

\bibitem{cocoapi}
T.-Y. Lin, P.~Doll{\'a}r, {\em et~al.}, ``{COCO API}.''
  \url{https://github.com/cocodataset/cocoapi}, Accessed on Apr. 14, 2023.

\bibitem{USB_Shinya_BMVC2022}
Y.~Shinya, ``{USB}: Universal-scale object detection benchmark,'' in {\em
  BMVC}, 2022.

\bibitem{GFL_NeurIPS2020}
X.~Li, W.~Wang, L.~Wu, S.~Chen, X.~Hu, J.~Li, J.~Tang, and J.~Yang,
  ``{Generalized Focal Loss}: Learning qualified and distributed bounding boxes
  for dense object detection,'' in {\em NeurIPS}, 2020.

\bibitem{nwd2}
C.~Xu, J.~Wang, W.~Yang, H.~Yu, L.~Yu, and G.-S. Xia, ``Detecting tiny objects
  in aerial images: A normalized wasserstein distance and a new benchmark,''
  {\em ISPRS Journal of Photogrammetry and Remote Sensing}, vol.~190,
  pp.~79--93, 2022.

\bibitem{zhang2021varifocalnet}
H.~Zhang, Y.~Wang, F.~Dayoub, and N.~Sunderhauf, ``Varifocalnet: An iou-aware
  dense object detector,'' in {\em CVPR}, 2021.

\bibitem{wang2021scaled}
C.-Y. Wang, A.~Bochkovskiy, and H.-Y.~M. Liao, ``Scaled-yolov4: Scaling cross
  stage partial network,'' in {\em CVPR}, 2021.

\bibitem{zhang2017mixup}
H.~Zhang, M.~Cisse, Y.~N. Dauphin, and D.~Lopez-Paz, ``mixup: Beyond empirical
  risk minimization,'' {\em arXiv}, 2017.

\bibitem{rekavandi2022guide}
A.~M. Rekavandi, L.~Xu, F.~Boussaid, A.-K. Seghouane, S.~Hoefs, and
  M.~Bennamoun, ``A guide to image and video based small object detection using
  deep learning: Case study of maritime surveillance,'' {\em arXiv}, 2022.

\bibitem{bosquet2021correlation}
B.~Bosquet, M.~Mucientes, and V.~M. Brea, ``Correlation-based convnet for small
  object detection in videos,'' in {\em ICPR}, 2021.

\bibitem{zhu2023tiny}
Y.~Zhu, C.~Li, Y.~Liu, X.~Wang, J.~Tang, B.~Luo, and Z.~Huang, ``Tiny object
  tracking: A large-scale dataset and a baseline,'' {\em Transactions on Neural
  Networks and Learning Systems}, 2023.

\bibitem{zhang2022tracking}
Z.~Zhang, F.~Wu, Y.~Qiu, J.~Liang, and S.~Li, ``Tracking small and fast moving
  objects: A benchmark,'' in {\em ACCV}, 2022.

\end{thebibliography}

\end{document}